\pdfoutput=1

\documentclass[11pt]{article}

\usepackage{acl}

\usepackage{times}
\usepackage{latexsym}

\usepackage[T1]{fontenc}

\usepackage[utf8]{inputenc}

\usepackage{microtype}

\usepackage{inconsolata}

\usepackage{amsmath}
\usepackage{amsfonts}
\usepackage{amssymb}
\DeclareMathOperator*{\argmax}{argmax}
\usepackage{bm}
\usepackage{algorithm}
\usepackage[noend]{algpseudocode}

\usepackage{multirow,multicol}
\usepackage{booktabs}
\usepackage{graphicx}
\usepackage{caption}
\usepackage{subfigure}

\newcommand{\methodname}{{\sc Self-Filter}}
\def\ie{$i.e.$}
\def\eg{$e.g.$}

\newcommand{\x}{\bm{x}}
\newcommand{\y}{\bm{y}}

\newcommand{\cL}{\mathcal{L}}
\newcommand{\cS}{\mathcal{S}}

\newcommand{\bbR}{\mathbb{R}}

\usepackage{hyperref}

\title{Your Vision-Language Model Itself Is a Strong Filter: Towards High-Quality Instruction Tuning with Data Selection}

\author{Ruibo Chen, Yihan Wu, Lichang Chen, Guodong Liu, Qi He \\ {\bf Tianyi Xiong},{\bf Chenxi Liu}, {\bf Junfeng Guo}, {\bf Heng Huang} \\
        University of Maryland, College Park \\
        \texttt{rbchen@umd.edu}
        }

\begin{document}

\maketitle
\begin{abstract}

Data selection in instruction tuning emerges as a pivotal process for acquiring high-quality data and training instruction-following large language models (LLMs), but it is still a new and unexplored research area for vision-language models (VLMs). Existing data selection approaches on LLMs either rely on single unreliable scores, or use downstream tasks for selection, which is time-consuming and can lead to potential over-fitting on the chosen evaluation datasets. To address this challenge, we introduce a novel dataset selection method, \methodname, that utilizes the VLM itself as a filter. This approach is inspired by the observation that VLMs benefit from training with the most challenging instructions. \methodname\ operates in two stages. In the first stage, we devise a scoring network to evaluate the difficulty of training instructions, which is co-trained with the VLM. In the second stage, we use the trained score net to measure the difficulty of each instruction, select the most challenging samples, and penalize similar samples to encourage diversity. Comprehensive experiments on LLaVA and MiniGPT-4 show that \methodname\ can reach better results compared to full data settings with merely about 15\% samples, and can achieve superior performance against competitive baselines. Our code is available at \href{https://github.com/RayRuiboChen/Self-Filter}{https://github.com/RayRuiboChen/Self-Filter}

\end{abstract}

\section{Introduction}


Instruction-following foundation models, such as Gemini~\cite{team2023gemini}, GPT-4 and GPT-4 Vision~\cite{openai2023gpt4}, have shown exceptional performance in multi-modality tasks. Their capabilities are primarily derived from the Reinforcement Learning with Human Feedback (RLHF)~\cite{ouyang2022training} paradigm and the Instruction Fine-Tuning (IFT)~\cite{longpre2023flan} framework. 
LIMA~\cite{zhou2023lima} indicates that while foundation models primarily acquire their knowledge during pre-training phases, only a small amount of instruction is needed for fine-tuning these models to produce the desired outputs during interactions. However, the low-quality instructions can significantly degrade the models' performance. To address this challenge, dataset selection tasks have been proposed, which aim to select high-quality instruction-tuning data to enhance the performance of instruction-following foundation models.

Existing data selection methods have many limitations in selecting high-quality instruction-tuning data. Traditional data pruning metrics~\citep{toneva2018empirical,meding2021trivial,yang2022dataset} usually use the influence on model accuracy to filter samples, \emph{i.e.}, dependent on the classification tasks, and are unsuitable for the generative models. \citet{zhou2023lima} manually crafted high-quality instruction dataset, which is costly and time-consuming. \citet{chen2023alpagasus} query ChatGPT to rank instructions, but the performance is unstable and there is no information from images in the multimodal setting. Other works~\citep{cao2023instruction,li2023one} use pre-defined evaluation datasets to construct metrics, thus highly relying on the quality and the distribution of the selected evaluation datasets, which can potentially cause over-fitting problems and harm the models' generalization ability. These limitations prevent existing data selection approaches from being applied to VLMs. To our best knowledge, InstructionGPT-4~\citep{wei2023instructiongpt} is the only dataset selection method focusing on VLMs. They also introduce additional tasks to train a data selector, and their experiments are only limited to MiniGPT-4 on a small dataset with 3.4k samples. 



To address the aforementioned issues, we propose \methodname\ by leveraging VLM itself as an effective data filter. Specifically, We first generate score embeddings for each training instruction through feature extractors, such as CLIP~\citep{radford2021learning} and GPT-4 Vision, with the training instruction itself as inputs. After obtaining score embeddings, we use them to train a score net together with the target VLM. The score net takes the score embeddings as input and produces the weight of corresponding training instructions in VLM training losses. We expect that these weights can indicate the ``difficulty" of the samples to be learned by VLMs during instruction tuning. Then we choose the instruction data with the highest difficulty to construct the filtered dataset. Moreover, we introduce a penalty mechanism on similar training samples to enhance the diversity of instructions. 
Compared with InstructionGPT-4~\citep{wei2023instructiongpt}, our method can easily transfer to any vision-language model with large-scale datasets.

Through comprehensive experiments on the widely studied vision-language models (\textit{e.g.,} LLaVA and MiniGPT-4), we show that with only around 15\% samples of the raw instruction tuning dataset, \methodname\ can surpass the model trained on the original instruction data across various evaluation datasets and benchmarks. \methodname\ also outperforms several competitive baselines and achieves state-of-the-art performance. 



Our contributions are summarized as follows:
\begin{itemize}
\vspace{-3pt}
    \item We propose a novel method named \methodname, and demonstrate that large vision-language models themselves can serve as filters for instruction-finetuning. We show the efficacy of our method through extensive experiments.
\vspace{-3pt}    \item To the best of our knowledge, we are the first to show that on large-scale instruction tuning datasets, vision-language models do not necessarily require a large number of data. A small amount of high-quality data is sufficient for successful instruction tuning. 
\vspace{-3pt}    \item Our method does not require additional pre-defined evaluation tasks or surrogate models. Importantly, it makes no assumptions about downstream tasks, thereby preserving the model's generalization capabilities—a critical factor for addressing complex real-world applications.

\end{itemize}

\section{Related Work}

\subsection{Data Selection}

Data selection is an emerging topic in large-language model instruction-tuning~\cite{chen2023alpagasus, cao2023instruction}, aiming at selecting high-quality data and discarding harmful data, which could cause hallucinations. Alpagasus~\citep{chen2023alpagasus} is the pioneer of automatic filtering. They query ChatGPT for instruction quality, improving the efficiency of training. \citet{li2023quantity} proposes using the IFD score as a proxy of the sample difficulty. Instruction Mining~\citep{cao2023instruction} uses a linear combination of several indicators to judge the sample quality. Concurrent work by \citet{li2023one} proposes to use one-shot learning performance on pre-defined tasks to filter data.

InstructionGPT-4~\citep{wei2023instructiongpt} is the only work in VLM instruction tuning. They utilize the combination of multimodal scores as the indicators and rely on a regression model trained on pre-defined tasks to conduct data selection. Their application is also limited to MiniGPT-4 with only 3.4k instructions.

\subsection{Instruction Finetuning}
Instruction tuning is an essential step for pre-trained foundation models to obtain instruction-following capabilities, and deal with complex human queries. 
Alpaca~\citep{taori2023alpaca}, Vicuna~\citep{chiang2023vicuna}, WizardLM~\citep{xu2023wizardlm} distill GPT-family models and release high-quality instruction-tuning datasets and build up the easy-to-use tools for instruction-finetuning.
Following them, MiniGPT-4~\citep{zhu2023minigpt} and LLaVA~\citep{liu2023visual} first utilize instruction tuning in multimodal models. They adopt the same pipeline, designing templates to distill GPT-family models and obtain data. After that, more vision-language models like InstructBlip~\cite{dai2023instructblip} and Qwen-VL-Chat~\cite{bai2023qwen} also exploit the instruction tuning procedure to gain chat ability.

A detailed related work discussion can be found in Appendix~\ref{sec:more_relatedwork}.

\begin{figure*}
    \centering
    \vspace{-0.5cm}
    \includegraphics[width=0.99\linewidth]{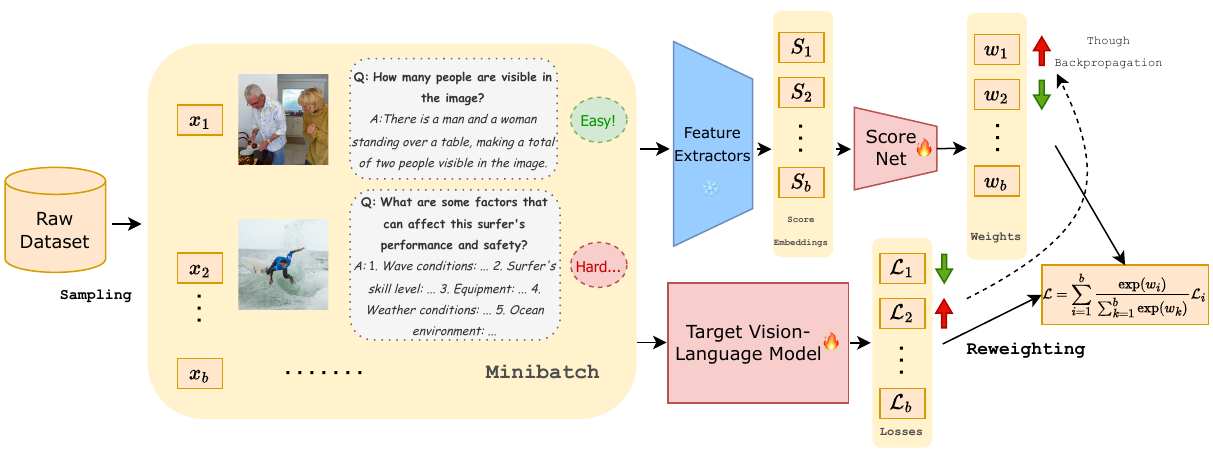}
    \vspace{-0.4cm}
    \caption{Illustration of the stage 1 of \methodname. 
    In the target vision-language models, samples that are more challenging usually yield higher losses. During the training process, the loss of each instruction is combined with a learnable weight generated by the score net. Through minimizing this weighted loss function $\mathcal{L}$, the instructions with higher loss tend to have lower weights. It is important to note that in stage 1, both the target VLM and the score net are actively trained, with the pre-trained feature extractors remaining frozen. Following this, stage 2 involves employing the score net to estimate the weights for each instruction, in which a lower weight signifies a higher difficulty level for the model.
}
\vspace{-0.4cm}
    \label{fig:arch}
\end{figure*}

\section{Methodology}

In this section, we briefly review the related background knowledge of vision-language models and elaborate on the task formulation. Then we thoroughly introduce our proposed \methodname.

\subsection{Preliminary}

\paragraph{Instruction-Following Vision-Language Models.}  The training process of current instruction-following vision-language models can roughly be divided into two parts, (1) \textit{pre-training}, (2) \textit{instruction fine-tuning (IFT)}, or \textit{reinforcement learning from human feedback (RLHF)}. During pre-training, the models may fine-tune their image and text backbones and align their representations. For IFT and RLHF, the models will gain instruction-following ability to output high-quality responses in conversations.

\paragraph{Task Formulation.} We present the definition of our data selection task in instruction-finetuning. Given the instruction tuning dataset $\mathcal{D}=\{\x_j\}_{j=1}^{N}$, where $\x_j=(\x_j^i,\x_j^t)$ denotes the input image and text pair, our task is to select a subset and to prune $\mathcal{D}$ to any desired size $m$, so that the vision-language model $f$ can achieve the best performance on downstream tasks $\{T_i\}_{i=1}^t$ after training on the selected subset $\mathcal{D}_{f}^{m}\subset \mathcal{D} $, $|\mathcal{D}_{f}^m|=m$.
Here the VLM $f$ is already pre-trained. We present the total VLM loss as $\cL$, and the VLM loss on a single instruction $\x_i$ as $\cL_i$.

Notably, we could not have access to the distribution of downstream tasks in real-world applications, so we should preserve the models' generalization ability while conducting data selection.

\subsection{SELF-FILTER}

\paragraph{Motivation.} The goal of data selection tasks is to find the instructions with the highest quality and diversity. In assessing the quality of instruction data, prior research~\cite{xu2023wizardlm, zhao2023preliminary,cao2023instruction} has indicated that the most challenging samples, or ``hardest" samples, can provide significant insights and are therefore valuable for model training. These ``hardest" samples are defined as those that present substantial learning challenges to VLMs during the training process. Motivated by this observation, we aim to develop an evaluation metric capable of quantifying the learning difficulty $d_i$ (with VLMs) of each sample $\x_i$ to accurately reflect its quality.

While many metric-based data selection methods~\cite{chen2023alpagasus,cao2023instruction,liu2023makes} have been applied to large language models (LLMs), directly transferring them to more complex multimodal contexts proves challenging, often yielding suboptimal outcomes as evidenced by our experiments. Given the difficulty of evaluating multimodal instructions with only a singular metric, we propose to combine these features via a lightweight scoring network $s$, which can be either a Multi-Layer Perceptron (MLP) or a linear layer.

To increase the diversity within the filtered instruction dataset, we introduce a penalty mechanism targeting similar examples. Specifically, when a sample is chosen for the filtered dataset, we reduce the likelihood of selecting its k-nearest neighbors from the raw dataset. This approach is designed to ensure a broader variety of instructional content by actively reducing redundancy among the chosen samples.

In the rest part of this section, we introduce \methodname, which consists of two stages: a) score net training, and b) data selecting by quality and diversity.

\begin{algorithm}[!t]
  \caption{Stage 1: Train the score net}
  \label{alg:stage1}
  \begin{algorithmic}
    \Require Raw multimodal instruction dataset $\mathcal{D}$, target vision-language model $f$, score net $s$ and its parameters $\phi$,  batch size $b$, feature extractors $M$
    \For {each step}
      \State Sample a minibatch $B=\{\x_i\}_{i=1}^{b} \sim \mathcal{D}$

      \For {$\x_i$ in $B$}
        \State Calculate the VLM loss $\mathcal{L}_i$ of $\x_i$
        \State  $\bm S_i \gets M(\x_i)$
        \State  $w_i \gets s(\bm S_i,\phi)$
      \EndFor
      \State $w_i' \gets \frac{\exp(w_i)}{\sum_{j=1}^{b}\exp(w_j)}b$
      \State  $\mathcal{L} \gets \frac{1}{b}\sum_{i=1}^{b} w_i' \mathcal{L}_i$
      \State Use $\mathcal{L}$ to backpropagate and optimize the 
      \State vision-language model and the score net
    \EndFor
  \end{algorithmic}
\end{algorithm}

\paragraph{Stage 1: Training the Score Net.} 

\citet{paul2021deep} identifies loss as an important indicator for measuring the difficulty level of training instructions, with more challenging instructions typically exhibiting higher loss. Motivated by data re-weighting approaches~\citep{gao2023self,zhang2021learning}, we propose utilizing learnable weights, produced by a score net, within the VLM loss framework to assess the difficulty of training instructions. Given that more challenging instructions typically incur higher loss, their corresponding learnable weights are inclined to decrease during the training process. Consequently, we can assess and rank the difficulty of instructions based on their learnable weights.
 The illustration of stage 1 of \methodname\ is shown in Figure~\ref{fig:arch}.

The initial phase in training the score net involves preparing the input. This is accomplished by extracting features from the training instructions using various extractors, such as CLIP and GPT-4V. Let $M=\{M_i\}_{i=1}^n$ denote the series of feature extractors. We combine the output features from these extractors to form the input of the score net $\bm S_i\in\cS$:
\begin{equation}
    \bm S_i=M(\x_i)=[M_1(\x_i),\cdots,M_n(\x_i)]
\end{equation}
After preparing the input, we use a score net $s(\cdot,\phi):\cS\to\bbR$, where $\phi$ represents the parameters of the score net, to project the generated input embeddings to the weight of the corresponding instructions in the learning objective of VLMs. Given a sample $\x_i$ and corresponding input embedding $\bm S_i$, the weight of $\x_i$ in the VLM loss is given by $w_i=s(\bm S_i,\phi)$. Denoted by $\cL_i$ the original VLM loss on $\x_i$, the new loss after reweighting is given by $w_i\cL_i$.

It is important to note that the weights generated by the score net for different instructions are independent and may vary significantly in scale, rendering them inadequate for cross-instruction difficulty ranking. To overcome this issue, we introduce an inter-batch weight normalization technique that applies the softmax function to the weights within the sampled batch. This results in a normalized weight, denoted as $w_i'=\frac{\exp(w_i)}{\sum_{k=1}^{b}\exp(w_k)}b$, where $b$ is the batch size. Our learning objective is subsequently formulated as follows:
\begin{equation}
    \mathcal{L} = \sum_{i=1}^{b} \frac{\exp(w_i)}{\sum_{k=1}^{b}\exp(w_k)} \mathcal{L}_i
\end{equation}

Through combining the weights with the VLM loss, the score net can receive guidance along with the training process of the VLM itself, avoiding the introduction of possibly unreliable surrogate models or additional evaluation datasets, and preserving the generalization ability of our target VLM. The detailed algorithm of stage 1 is shown in Algorithm~\ref{alg:stage1}.

In our experiments, we observe that utilizing a limited set of pre-calculated scores as feature extractors $M$, \eg, CLIP Score~\citep{hessel2021clipscore} and GPT Score~\citep{chen2023alpagasus}, yields satisfactory results. However, for larger data volumes, more comprehensive features like CLIP-encoded features tend to outperform, as they contain richer information. This enables the score net $s$ to more effectively find the relationship between coarse features and instruction quality through larger scale training.

\begin{algorithm}[!htp]
  \caption{Stage2: Filter the instructions}
  \label{alg:stage2}
  \begin{algorithmic}
    \Require Raw multimodal instruction finetuning dataset $\mathcal{D}$, target pruning size $m$, k-nearest neighbor algorithm \textit{KNN}, similarity metric \textit{Sim}
    \State $\mathcal{D}_f^m \gets \{\}$
    \For {$\x_i$ in $\mathcal{D}$} 
        \State $\bm S_i \gets M(\x_i)$
        \State $w_i \gets s(\bm S_i,\phi)$ 
        \State $d_i \gets -w_i$
    \EndFor
    \For{$m$ iterations}
        \State $i \leftarrow \argmax_{i,\x_i \in \mathcal{D}} \ d_i$
        \State $\mathcal{D}_f^m \gets \mathcal{D}_f^m \cup \{\x_i\}$ 
        \State $\mathcal{D} \gets \mathcal{D} - \{\x_i\}$
        \For{ $\x_j$ in $\text{KNN}(\x_i)$} 
            \State $d_j \gets d_j-\gamma \text{Sim}(\x_i,\x_j)^2 d_i$
        \EndFor
    \EndFor
    \State \textbf{Output} $\mathcal{D}_f^m$
  \end{algorithmic}
\end{algorithm}

\paragraph{Stage 2: Filtering the Instructions Considering Quality and Diversity.}

After obtaining the trained score net $s$ in stage 1, we proceed to compute the final weight $w_i$ for each sample. Intuitively, more challenging multimodal instructions produce larger losses during training, which in turn, through backpropagation, results in a smaller weight $w_i$. Therefore, we define our difficulty metric as $d_i=-w_i$.

We do not directly use the training loss to rank instructions due to the observation that newly trained samples typically exhibit smaller losses. This could lead to biased ranking and, as demonstrated in our experiments (see Section~\ref{sec:results}), suboptimal performance.

Moreover, diversity has been recognized as a critical factor for effective data selection~\citep{wei2023instructiongpt,maharana2023d2}. In our approach, we enhance diversity by penalizing the k-nearest neighbors of a newly selected sample, explicitly lowering their difficulty scores to encourage a broader variety of selections. After selecting an instruction $\x_i$ to the filtered dataset, we adjust the difficulty of its k-nearest neighbors examples by:
\begin{equation}
    d_j=d_j-\gamma \text{Sim}(\x_i,\x_j)^2 d_i\,,
    \label{eq:diversity}
\end{equation}
where $\x_j$ is one of the k-nearest neighbors of $\x_i$, and $\text{Sim}(\x_i,\x_j) \in \mathbb{R}$ represents their similarity. $\gamma$ is a hyperparameter and is empirically set to 1. The whole process of stage 2 is described in Algorithm~\ref{alg:stage2}.

\section{Experiments}

\begin{table*}[!t]
\caption{\footnotesize Main results for \methodname\ on LLaVA~\citep{liu2023visual}. \textit{Scores} represents using CLIP Score, Imagereward and ChatGPT as feature extractors, and \textit{CLIP} presents using CLIP features. We also try removing the diversity module under the \textit{Scores} setting. The meanings of the numbers are detailed in Appendix~\ref{sec:eval}.}
\label{tab:llava_main} 
\centering
\renewcommand\arraystretch{1.15}
\setlength{\tabcolsep}{1.4mm}{}
\small
\begin{tabular}{cc|cccccc|ccc}
\toprule
 &                     & \multicolumn{6}{c|}{\textbf{Baseline}}                     & \multicolumn{3}{c}{\textbf{\methodname}}                                                                      \\ \midrule
\textbf{}                                                                                                & \textbf{}           & \multicolumn{1}{c|}{\textbf{Full data}} & \textbf{Random} & \textbf{GraNd} & \textbf{EL2N} & \textbf{\begin{tabular}[c]{@{}c@{}}Prototy-\\ picality\end{tabular}} & \textbf{\begin{tabular}[c]{@{}c@{}}Alpaga-\\ sus\end{tabular}} & \textbf{Scores} & \textbf{\begin{tabular}[c]{@{}c@{}}w/o \\ Diversity\end{tabular}} & \textbf{CLIP}    \\ \midrule
\multicolumn{2}{c|}{\textbf{Samples}}                                                                                          & \multicolumn{1}{c|}{158k}               & 25k             & 25k            & 25k           & 25k                                                                  & 25k                                                            & 25k             & 25k                                                               & 25k              \\ \midrule
\multicolumn{2}{c|}{\textbf{MMBench}}                                                                                          & \multicolumn{1}{c|}{23.97}              & 30.67           & 28.26          & 30.41         & 29.64                                                                & 34.71                                                          & 30.58           & 24.66                                                             & \textbf{38.48}   \\ \cline{1-2}
\multicolumn{1}{c|}{\multirow{3}{*}{\textbf{MME}}}                                                       & \textbf{Overall}    & \multicolumn{1}{c|}{1132.89}            & 1093.15         & 1138.49        & 969.48        & 1165.69                                                              & 1096.05                                                        & 1126.24         & 1079.9                                                            & \textbf{1218.15} \\
\multicolumn{1}{c|}{}                                                                                    & \textbf{Perception} & \multicolumn{1}{c|}{884.68}             & 858.15          & 884.49         & 723.77        & 904.98                                                               & 827.84                                                         & 850.89          & 804.19                                                            & \textbf{955.65}  \\
\multicolumn{1}{c|}{}                                                                                    & \textbf{Reasoning}  & \multicolumn{1}{c|}{248.21}             & 235.00          & 254.00         & 245.71        & 260.71                                                               & 268.21                                                         & 275.35          & \textbf{275.71}                                                   & 262.50           \\ \cline{1-2}
\multicolumn{2}{c|}{\textbf{SEED-Bench}}                                                                                        & \multicolumn{1}{c|}{39.96}              & 44.38           & 45.15          & 43.86         & 46.80                                                                & \textbf{47.67}                                                 & 45.34           & 40.94                                                             & 47.54            \\ \cline{1-2}
\multicolumn{1}{c|}{\multirow{3}{*}{\textbf{\begin{tabular}[c]{@{}c@{}}Hallusion\\ Bench\end{tabular}}}} & \textbf{aACC}       & \multicolumn{1}{c|}{45.32}              & 43.85           & 42.27          & 41.22         & 45.32                                                                & 42.69                                                          & \textbf{48.15}  & 45.53                                                             & 45.85            \\
\multicolumn{1}{c|}{}                                                                                    & \textbf{fACC}       & \multicolumn{1}{c|}{\textbf{14.74}}     & 14.16           & 9.83           & 9.25          & 13.29                                                                & 10.12                                                          & 11.54           & 10.98                                                             & 14.45            \\
\multicolumn{1}{c|}{}                                                                                    & \textbf{qACC}       & \multicolumn{1}{c|}{10.33}              & 9.01            & 8.57           & 7.25          & 10.77                                                                & 6.81                                                           & \textbf{11.11}  & 10.99                                                             & 10.33            \\ \cline{1-2}
\multicolumn{2}{c|}{\textbf{MathVista}}                                                                                        & \multicolumn{1}{c|}{25.10}              & 24.10           & 24.60          & 24.80         & 25.40                                                                & 23.90                                                          & 25.40           & \textbf{26.70}                                                    & 26.40            \\ \cline{1-2}
\multicolumn{2}{c|}{\textbf{ScienceQA}}                                                                                        & \multicolumn{1}{c|}{55.03}              & 56.12           & 52.79          & 58.01         & 56.87                                                                & 58.75                                                          & 56.22           & 55.83                                                             & \textbf{59.40}   \\ \bottomrule
\end{tabular}
\end{table*}

\begin{figure*}[!htb]
    \centering
    \includegraphics[width=0.99\linewidth]{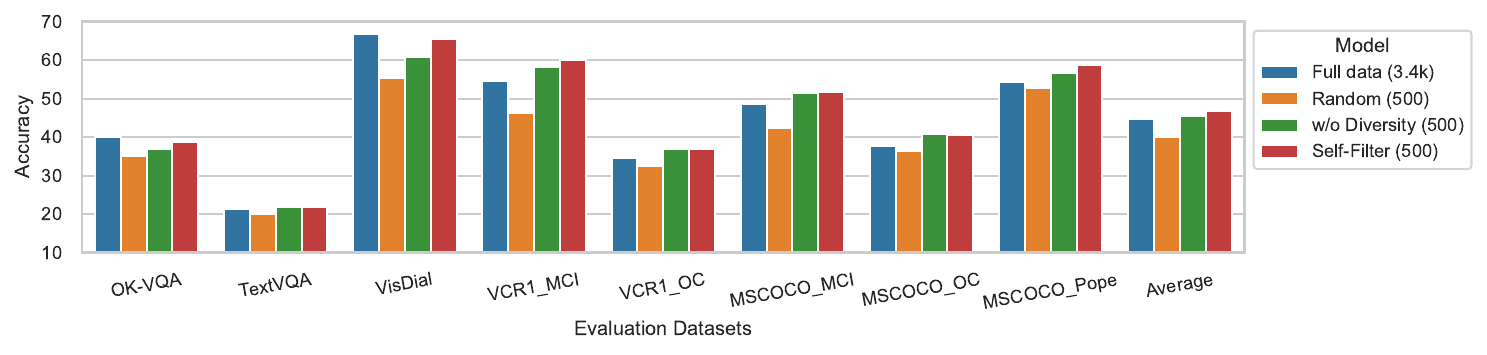}
    \vspace{-0.3cm}
    \caption{Main results on MiniGPT-4~\citep{zhu2023minigpt}. \textit{w/o Diversity} represents removing the diversity module in our method. \methodname\ achieves the best results on most tasks, and has a higher average accuracy.}
    \label{fig:minigpt4_main}
    \vspace{-0.3cm}
\end{figure*}

In this section, we first detail our settings and the chosen hyperparameters. Then we introduce the evaluation benchmarks and datasets used in our experiments and the baseline methods. Finally, we show that our proposed \methodname\ can get state-of-the-art performance through extensive experiments.

\subsection{Settings and Hyperparameters}

We evaluate our approach on two extensively studied vision-language models: LLaVA~\citep{liu2023visual} and MiniGPT-4~\citep{zhu2023minigpt}. These models employ vision encoders and large language models as backbones, and they design different pre-training tasks to train projection modules for aligning multimodal representations. Then they employ automatically generated multimodal instruction datasets through instruction-finetuning. Further information regarding the specific versions of the models is available in Appendix~\ref{sec:vlm_version}.

The instruction-tuning dataset in LLaVA comprises 157,712 training samples, whereas MiniGPT-4 has 3,439 samples. Our approach adheres strictly to the original instruction-tuning settings and hyperparameters as outlined in their official implementations. We use 8 A6000 GPUs in our experiments.

For the selection of feature extractors $M$, in MiniGPT-4, we incorporate CLIP Score~\citep{hessel2021clipscore}, Imagereward~\citep{xu2023imagereward}, ChatGPT, and GPT-4Vision. In LLaVA, the chosen feature extractors include CLIP Score, Imagereward, and ChatGPT (referred to as \textit{Scores} in Table~\ref{tab:llava_main}). Additionally, we experiment with using the features directly generated by the CLIP encoders (referred to as \textit{CLIP} in Table~\ref{tab:llava_main}). Comprehensive details about all the feature extractors and their utilization can be found in Appendix~\ref{sec:feat}.

For the score net, we implement it as a linear layer. We set the value of $k$ to 10 for the KNN algorithm in the diversity module and employ cosine similarity between CLIP features to measure the sample similarity, as described in Eq.~\ref{eq:diversity}.

\subsection{Evaluations}

\paragraph{Benchmarks and Datasets} We employ two evaluation tools, LVLM-eHub~\citep{xu2023lvlm} and VLMEvalKit~\citep{2023opencompass}, to conduct a comprehensive assessment of our method. Our evaluation encompasses a diverse array of benchmarks and datasets, including MMBench~\citep{liu2023mmbench}, MME~\citep{fu2023mme}, SEED-Bench~\citep{li2023seed}, HallusionBench~\citep{liu2023hallusionbench}, MathVista~\citep{lu2023mathvista}, ScienceQA~\citep{lu2022learn}, OK-VQA~\citep{marino2019ok}, TextVQA~\citep{singh2019towards}, VisDial~\citep{das2017visual}, VCR~\citep{zellers2019vcr}, MSCOCO~\citep{lin2014microsoft}, and Pope~\citep{li2023evaluating}. The interpretation of their evaluation metrics is detailed in Appendix~\ref{sec:eval}.

\paragraph{Baselines.}

We compare the performance of \methodname\ with random filtering and four other competitive baselines:

\begin{itemize}
    \item \textbf{EL2N:}~\citep{paul2021deep} 
    Initially introduced for classification tasks, EL2N (Error L2-Norm) score is defined as $\mathbb{E}\lVert p(\bm w,\x)-\y \lVert_2$, where $\bm w$ represents the parameters, $\x$ denotes the input features, and $\y \in \{0,1\}^K$ is the ground truth label in the form of a one-hot vector. To align with the generative model, we modify the score as the form of the average cross-entropy loss for a single instruction.

    \item \textbf{GraNd:}~\citep{paul2021deep} 
    GraNd (Gradient Normed) score is defined as $\mathbb{E}_{\bm{w}} \lVert g(\x,\y) \rVert_2$, which utilizes the norm of gradients caused by each sample to quantify their influence. A higher GraNd score suggests that the corresponding sample is more important.

    \item \textbf{Prototypicality:}~\citep{sorscher2022beyond} The prototypicality metric involves performing k-means clustering in the embedding space. The difficulty of each data point is determined by the Euclidean distance to its nearest cluster centroid, \ie, the prototype. In our experiments, we utilize the last hidden layer feature as the embedding.

    \item \textbf{Alpagasus:}~\citep{chen2023alpagasus} 
    Alpagasus queries ChatGPT to provide evaluation scores and detailed explanations for each data point. Then these scores are utilized for ranking and filtering instructions.

\end{itemize}

Our method, along with all the aforementioned baselines, does not require additional datasets or surrogate models, ensuring a fair comparison.

\begin{table*}[!htb]
\tiny
\renewcommand\arraystretch{1.17}
\caption{The easiest sample and hardest sample produced by \methodname.}
\label{tab:easiest_and_hardest}
\centering
\begin{tabular}{l|l|l}
\toprule
\textbf{\begin{tabular}[c]{@{}l@{}}Easiest\\ Sample\end{tabular}}  & \multicolumn{1}{c|}{\textbf{Text}}   & \multicolumn{1}{c}{\textbf{Image}} \\ \midrule
\textbf{{[}Human{]}:}                                              & How many people are visible in the image?  & \multirow{6}{*}{\includegraphics[width=0.06\linewidth]{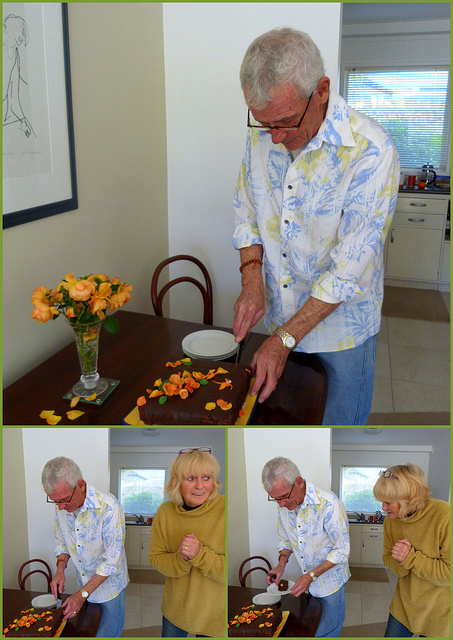}}                  \\
\textbf{{[}GPT{]}:}                                                & There is a man and a woman standing over a table, making a total of two people visible in the image.  &                                    \\
\textbf{{[}Human{]}:}                                              & What is the man doing in the image?           &                                    \\
\textbf{{[}GPT{]}:}                                                & In the image, the man is cutting into a chocolate frosted cake. &                                    \\
\textbf{{[}Human{]}:}                                              & Are there any other elements on the table besides the cake?                                                           &                                    \\
\textbf{{[}GPT{]}:}                                                & Yes, there are orange flowers on the table in addition to the cake.  &                                    \\ \midrule
\textbf{\begin{tabular}[c]{@{}l@{}}Hardest \\ Sample\end{tabular}} & \multicolumn{1}{c|}{\textbf{Text}}   & \multicolumn{1}{c}{\textbf{Image}} \\ \midrule
\textbf{{[}Human{]}:}                                              & What precautions should someone take when planning to surf in rough waters?& \multirow{2}{*}{ \includegraphics[width=0.08\linewidth]{figures/000000129758.jpg}}                  \\
\textbf{{[}GPT{]}:}                                                & \begin{tabular}[c]{@{}l@{}}When planning to surf in rough waters, like the man wearing a black wetsuit in the image, it is essential to take several precautions to ensure \\safety. These precautions include:\\ 1. Wearing appropriate gear: A wetsuit provides buoyancy, warmth, and protection from any potential hazards in the water, such as rocks.\\ 2. Checking weather and water conditions: Understanding the current ocean conditions and the forecast will help surfers anticipate how rough \\the waves might be and decide if it's safe to surf.\\ 3. Being aware of rip currents: These strong currents can carry surfers away from the shore, so it's crucial to understand how to spot and \\escape them.\\ 4. Surfing with a buddy: Having a surfing partner ensures there's someone to help or alert others in case of an emergency.\\ 5. Knowing personal limits: Understanding one's skill level and not attempting to surf waves that are beyond that level is essential for safety.\\ 6. Proper warm-up and stretching: This can help prevent injuries while surfing in challenging conditions.\\ 7. Communicating with local surfers: They can offer insights into the specific beach break and advise on potential hazards and safe spots.\\ By taking these precautions, surfers can significantly reduce the risks associated with surfing in rough waters and enjoy the exhilaration and \\challenges that such conditions offer.\end{tabular} &                                    \\ \bottomrule
\end{tabular}
\end{table*}

\subsection{Experimental Results}
\label{sec:results}

\paragraph{MiniGPT-4 Results.}

We evaluate MiniGPT-4 with LVLM-eHub, and the results are presented in Figure~\ref{fig:minigpt4_main}. With only 500 samples (14.5\% of the dataset), \methodname\ achieves the best results in 6 out of 8 tasks, significantly outperforming the random baseline. Additionally, our findings demonstrate that the diversity module contributes to enhancing the selection process. The specific numerical results for MiniGPT-4 can be found in Appendix~\ref{sec:minigpt4_number}.

\begin{table}[!htb]
\caption{Results for selecting the hardest samples and the easiest samples. 25k samples are used in training. Choosing the easiest samples found by \methodname\ can greatly hurt the performance.}
\label{tab:easiest}
\small
\centering
\renewcommand\arraystretch{1.15}
\setlength{\tabcolsep}{0.8mm}{}
\begin{tabular}{cc|cc}
\toprule
&            & \textbf{Hardest (Ours)} & \textbf{Easiest} \\ \midrule
\multicolumn{2}{c|}{\textbf{MMBench}}                                       & \textbf{38.48}          & 24.14            \\ \cline{1-2}
\multicolumn{1}{c|}{\multirow{3}{*}{\textbf{MME}}}             & \textbf{Overall}    & \textbf{1218.15}        & 962.36           \\ \cline{2-2}
\multicolumn{1}{c|}{}                                          & \textbf{Perception} & \textbf{955.65}         & 713.43           \\ \cline{2-2}
\multicolumn{1}{c|}{}                                          & \textbf{Reasoning}  & \textbf{262.5}          & 248.93           \\ \cline{1-2}
\multicolumn{2}{c|}{\textbf{SEED-Bench}}                                     & \textbf{47.54}          & 43.18            \\ \cline{1-2}
\multicolumn{1}{c|}{\multirow{3}{*}{\textbf{Hallusion Bench}}} & \textbf{aACC}       & \textbf{48.15}          & 43.01            \\
\multicolumn{1}{c|}{}                                          & \textbf{fACC}       & \textbf{14.45}          & 12.14            \\
\multicolumn{1}{c|}{}                                          & \textbf{aACC}       & \textbf{10.33}          & 8.13             \\ \cline{1-2}
\multicolumn{2}{c|}{\textbf{MathVista}}                                     & \textbf{26.4}           & 25.1             \\ \cline{1-2}
\multicolumn{2}{c|}{\textbf{ScienceQA}}                                     & \textbf{59.4}           & 55.88            \\ \bottomrule
\end{tabular}
\vspace{-0.3cm}
\end{table}

\paragraph{LLaVA Results.}
Our main experimental results on LLaVA using VLMEvalKit are displayed in Table~\ref{tab:llava_main}. We train the model with 25k samples (15.9\%) selected from the original dataset of 158k instances, resulting in substantial improvements compared to the full data settings. While several competitive baseline methods can also surpass the full dataset performance, our approach leads to higher results.

Furthermore, we observe that using CLIP features (referred to as the \textit{CLIP} setting) leads to further performance improvement. This finding suggests that the score net can effectively extract information from coarse features. In contrast, pre-calculated scores such as CLIP Score and ChatGPT outputs may be prone to inaccuracies and information losses.

\section{Analysis}

In this section, we first show the importance of the diversity module and choosing harder samples through ablation studies. Then we show that our method can indeed select the difficult samples in the case study. Finally, we analyze the influence of pruning size and batch size.

\subsection{Ablation Study}

As Table~\ref{tab:llava_main} and Figure~\ref{fig:minigpt4_main} show, removing the diversity module and not imposing penalties on similar samples may maintain similar performance on certain datasets, but lead to degradation on others, such as VisDial for MiniGPT-4 and MMBench for LLaVA.

This underscores the importance of ensuring diversity during data selection. Given the potential variation in data quality across different topics, a sole emphasis on quality might result in an unbalanced distribution of topics.

\subsection{Why Choose the Hardest Samples?}

We try to select the easiest samples as the filtered dataset in LLaVA under the \textit{CLIP} setting, and the results are presented in Table~\ref{tab:easiest}. The evaluation scores experience a significant drop compared to the hardest policy, reaffirming the observations made by~\citet{xu2023wizardlm} that difficult samples are more informative and effective for instruction tuning.

\subsection{Case Study}

In this section, we investigate the samples selected and abandoned by our method. Table~\ref{tab:easiest_and_hardest} showcases the easiest and hardest samples identified by the score net from the training data in LLaVA. Additional examples can be found in Appendix~\ref{sec:casestudy}.

The discarded easy examples typically consist of uncomplicated and direct questions, such as counting the objects in an image and describing the scenario. In contrast, more challenging instances focus on intricate tasks involving reasoning and planning, often having greater length. The Pearson correlation between the difficulty score and the sample text length is -0.6325, reflecting the significance of length as a factor but also emphasizing that it cannot be the sole determinant of difficulty. In summary, our approach adeptly and precisely identifies challenging samples for instruction tuning.

Additionally, it is evident that the distribution of topics among difficulty levels is uneven, with the top 5 most challenging samples all talking about surfing. Consequently, it is necessary to design the diversity module to ensure a more balanced representation across various subjects.

\begin{figure}[t]
    \centering
    \subfigure[MMBench]{
    \includegraphics[width=0.9\linewidth]{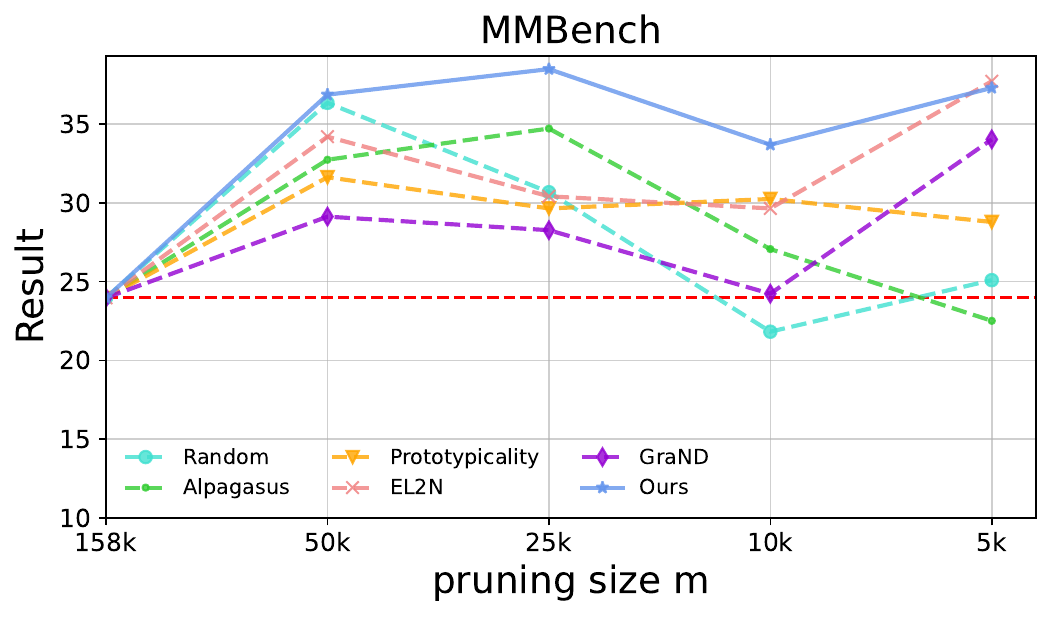}
    }
    \quad
    \subfigure[ScienceQA]{
    \includegraphics[width=0.9\linewidth]{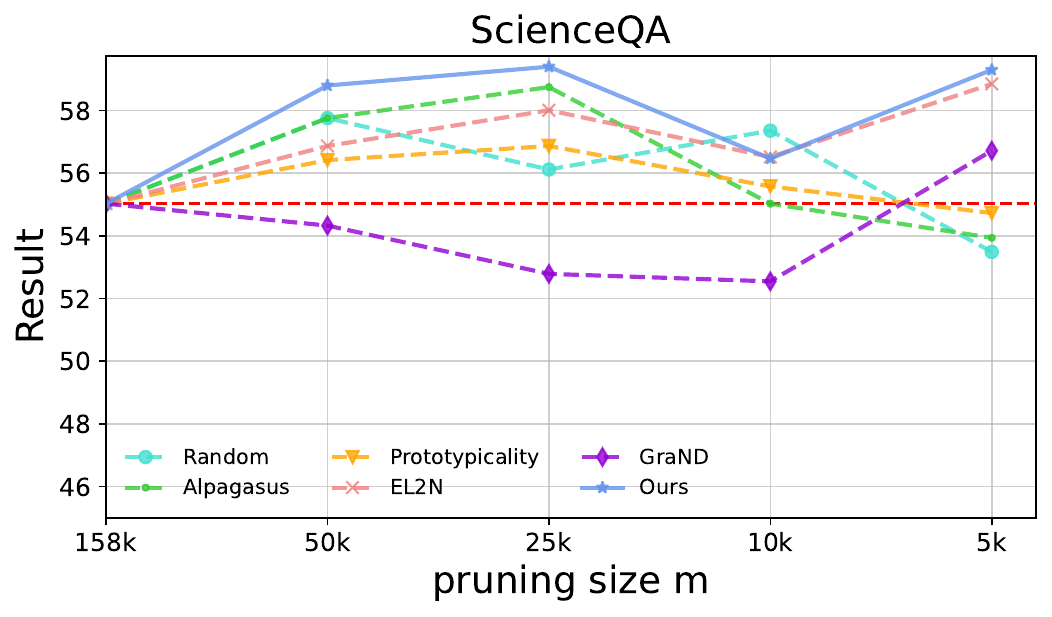}
    }
    \vspace{-0.3cm}
    \caption{Experimental results under difference target pruning size $m$.}
    
    \label{fig:datasize}
    \vspace{-0.3cm}
\end{figure}

\subsection{Performance under Different Pruning Size}

We conducted experiments with different target pruning sizes, and the results are illustrated in Figure~\ref{fig:datasize}. We use the \textit{CLIP} setting in LLaVA, and our method consistently outperforms others in the majority of cases. As the pruning size $m$ decreases, the evaluation results initially show improvement by discarding low-quality data, reaching a peak at 25k samples. Subsequently, the results gradually decline, emphasizing the prominence of quantity as a determining factor. It's noteworthy that limited data significantly impacts data diversity.

In traditional classification tasks, \citet{paul2021deep} observed a substantial performance decline with approximately 30\%-50\% data. In contrast, in the instruction tuning task, a mere 5k samples (3.17\%) still yield reasonable results, implying that little knowledge is learned during this stage.

\subsection{Influence of Different Batch Size}

We performed experiments using varying batch sizes during the training of the score net in Algorithm~\ref{alg:stage1}. The local batch size $b$ determines the number of observed samples when adjusting the weights. It's important to note that we concurrently modified the gradient accumulation steps to maintain the overall batch size for the vision-language model unchanged.

The results are presented in Table~\ref{tab:bsz}. In our primary experiments, we adhere to the LLaVA settings, employing a batch size of 16. Notably, the performance exhibits minimal variance across different local batch sizes, indicating the robustness of \methodname, capable of accurately extracting information even from a local window as small as 4 samples.

\begin{table}[t]
\small
\renewcommand\arraystretch{1.3}
\setlength{\tabcolsep}{0.8mm}{}
\centering
\caption{Results for different batch size $b$.}
\label{tab:bsz}
\begin{tabular}{cc|ccc}
\toprule
\multicolumn{2}{c|}{\textbf{Batch Size}} & \textbf{16} & \textbf{8} & \textbf{4} \\ \midrule
\multicolumn{2}{c|}{\textbf{MMBench}} & 38.48 & 37.46 & \textbf{43.13} \\ \cline{1-2}
\multicolumn{1}{c|}{\multirow{3}{*}{\textbf{MME}}} & \textbf{Overall} & \textbf{1218.15} & 1178.02 & 1179.24 \\
\multicolumn{1}{c|}{} & \textbf{Perception} & \textbf{955.65} & 930.52 & 924.95 \\
\multicolumn{1}{c|}{} & \textbf{Reasoning} & \textbf{262.5} & 247.5 & 254.29 \\ \cline{1-2}
\multicolumn{2}{c|}{\textbf{SEEDBench}} & 47.54 & \textbf{48.33} & 48.20 \\ \cline{1-2}
\multicolumn{1}{c|}{\multirow{3}{*}{\textbf{\begin{tabular}[c]{@{}c@{}}Hallusion\\ Bench\end{tabular}}}} & \textbf{aACC} & \textbf{45.85} & 45.22 & 45.01 \\
\multicolumn{1}{c|}{} & \textbf{fACC} & 14.45 & \textbf{15.03} & 13.01 \\
\multicolumn{1}{c|}{} & \textbf{qACC} & 10.33 & 10.11 & \textbf{10.77} \\ \cline{1-2}
\multicolumn{2}{c|}{\textbf{MathVista}} & \textbf{26.4} & 25.5 & 25.6 \\ \cline{1-2}
\multicolumn{2}{c|}{\textbf{ScienceQA}} & 59.4 & 58.55 & \textbf{60.09} \\ \bottomrule
\end{tabular}
\vspace{-0.3cm}
\end{table}


\section{Conclusion}

In this paper, we propose \methodname\ to select high-quality instructions from the noisy dataset. Our method leverages the vision-language model itself as a filter, and does not need any supplemental evaluation datasets, enhancing the model's generalization ability. We design a score net framework to effectively rank all the samples based on the predicted difficulty and discard low-quality instructions to help representation learning. Our comprehensive experiments on LLaVA and MiniGPT-4 demonstrate the effectiveness of our approach, and verify that large vision-language models learn most of their knowledge from the backbones and the pre-training stage, while a small amount of high-quality samples is sufficient for instruction tuning.

\clearpage

\section*{Limitations}

Though we prove the effectiveness of \methodname\ on LLaVA and MiniGPT-4, we could further extend the experiments on other vision-language models. A more careful search for hyperparameters, e.g., the batch size and $\gamma$ in the diversity module can generate better results.

Another limitation is that we could explore different architectures of the score net, for example, a small vision-language model. We could drop the feature extractors under this setting, and directly feed the samples into the score net to predict weights.

\bibliography{custom}

\clearpage
\appendix

\section{Additional Related Work}
\label{sec:more_relatedwork}
\subsection{Vision-Language Model}
Vision-language models have various training strategies and various architectures. CLIP~\citep{radford2021learning} uses contrastive learning to acquire high-quality unimodal representations. BLIP~\citep{li2022blip} designs a Q-Former to bridge two modalities.  Current instruction-following vision-language models, \eg, LLaVA~\citep{liu2023visual}, MiniGPT-4~\citep{zhu2023minigpt} and InstructBLIP~\citep{dai2023instructblip} usually exploit pre-trained language models (e.g., Llama~\citep{touvron2023llama1}, Vicuna~\citep{chiang2023vicuna}) and vision models (e.g., ViT~\citep{dosovitskiy2020image}, CLIP encoders) as backbones, and use projection modules to transform visual information into the embedding space of language models.

\subsection{Data Selection}
Data selection, or data pruning is always an important topic in the machine learning area. It tries to train a better model with less data, or greatly reduce the data size while do not lose too much model performance. \citet{toneva2018empirical} design a forgetting score to reflect whether the sample is easy to learn. \citet{paul2021deep} point out that loss and gradient norm can be informative metrics for pruning. \citet{sorscher2022beyond} choose to use the distance from the cluster centroid to measure difficulty.
\citet{meding2021trivial} propose to use multiple models to evaluate each image and classify it into ``trivial", ``impossible" or normal. \citet{yang2022dataset} examine the influence of the model's generalization ability when removing a sample. 

Data selection is an emerging topic in large-language model instruction-tuning~\cite{chen2023alpagasus, cao2023instruction}, aiming at selecting high-quality data and discarding harmful data, which could cause hallucinations. Alpagasus~\citep{chen2023alpagasus} is the pioneer of automatic filtering. They query ChatGPT for instruction quality, improving the efficiency of training. \citet{li2023quantity} proposes using the IFD score as a proxy of the sample difficulty. Instruction Mining~\citep{cao2023instruction} uses a linear combination of several indicators to judge the sample quality. Concurrent work by \citet{li2023one} proposes to use one-shot learning performance on pre-defined tasks to filter data.

A significant difference is that most previous methods~\citep{cao2023instruction,wei2023instructiongpt,li2023one} need additional evaluation tasks to learn the weights for different indicators, but we design a score net to automatically combine them, and our score net can utilize very coarse features like the CLIP feature which is a 1536-dimensional vector, avoiding too much information loss when calculating the indicators.

InstructionGPT-4~\citep{wei2023instructiongpt} is the only work in VLM instruction tuning. They utilize the combination of multimodal scores as the indicators and rely on a regression model trained on pre-defined tasks to conduct data selection. Their application is also limited to MiniGPT-4 with only 3.4k instructions.

\subsection{Instruction Finetuning}
Instruction tuning is an essential step for pre-trained foundation models to obtain instruction-following capabilities, and deal with complex human queries. 
Alpaca~\citep{taori2023alpaca}, Vicuna~\citep{chiang2023vicuna}, WizardLM~\citep{xu2023wizardlm} distill GPT-family models and release high-quality instruction-tuning datasets and build up the easy-to-use tools for instruction-finetuning.
Following them, MiniGPT-4~\citep{zhu2023minigpt} and LLaVA~\citep{liu2023visual} first utilize instruction tuning in multimodal models. They adopt the same pipeline, designing templates to distill GPT-4 and obtain data. After that, more vision-language models like InstructBlip~\cite{dai2023instructblip} and Qwen-VL-Chat~\cite{bai2023qwen} also exploit the instruction tuning procedure to gain chat ability.

\section{Vision-Language Model Version}
\label{sec:vlm_version}

For LLaVA, we use the LLaVA v1.0 7B model. The language backbone is Vicuna-7B-v1.3, and the vision encoder is clip-vit-large-patch14. For MiniGPT-4, we use Vicuna V0 7B as the language model.

\section{Feature Extractors}

\label{sec:feat}

\subsection{GPT Input Templates}

The data structures of the instruction tuning datasets for LLaVA and MiniGPT have some differences, so we design different GPT templates for them in Table~\ref{tab:gpt_templates}. The templates are adapted from \citet{chen2023alpagasus}.

We then use the output numbers as the scores for each instruction.

\begin{table*}[!htb]
\caption{GPT input templates.}
\label{tab:gpt_templates}
\centering
\small
\renewcommand\arraystretch{1.3}
\begin{tabular}{ll|l}
\toprule
\multicolumn{2}{c|}{\textbf{Model}}                                             & \multicolumn{1}{c}{\textbf{Template}}           \\ \midrule
\multicolumn{1}{l|}{\multirow{2}{*}{\textbf{MiniGPT-4}}} & \textbf{ChatGPT}     & \begin{tabular}[c]{@{}l@{}}We would like to request your feedback on the performance of AI assistant in \\ response to the instruction and the given input displayed following.\\ \\ Instruction: Describe this image in detail.\\ Response: {[}Caption{]}\\ \\ Please rate according to the quality of the response to the instruction. Each \\ assistant receives a score on a scale of 0 to 10, where a higher score indicates \\ higher level of the quality. Please first output a single line containing the value \\ indicating the scores. In the subsequent line, please provide a comprehensive \\ explanation of your evaluation, avoiding any potential bias.\end{tabular} \\ \cline{2-3} 
\multicolumn{1}{l|}{}                                    & \textbf{GPT-4Vision} & \begin{tabular}[c]{@{}l@{}}We would like to request your feedback on the performance of AI assistant in \\ response to the instruction and the given image above.\\ \\ Instruction: Describe this image in detail.\\ Response: {[}Caption{]}\\ \\ Please rate according to the quality of the response and the input image. Each \\ assistant receives a score on a scale of 0 to 10, where a higher score indicates \\ higher level of the quality. Please first output a single line containing the value \\ indicating the scores. In the subsequent line, please provide a comprehensive \\ explanation of your evaluation, avoiding any potential bias.\end{tabular}              \\ \midrule
\multicolumn{1}{l|}{\textbf{LLaVA}}                      & \textbf{ChatGPT}     & \begin{tabular}[c]{@{}l@{}}We would like to request your feedback on the performance of AI assistant in \\ response to the instruction and the given input displayed following.\\ \\ Conversation: {[}Conversation{]}\\ \\ Please rate according to the quality of the response to the instruction. Each \\ assistant receives a score on a scale of 0 to 10, where a higher score indicates \\ higher level of the quality. Please first output a single line containing the value \\ indicating the scores. In the subsequent line, please provide a comprehensive \\ explanation of your evaluation, avoiding any potential bias.\end{tabular}                                      \\ \bottomrule
\end{tabular}
\end{table*}

\subsection{Imagereward}

Imagereward~\citep{xu2023imagereward} is a general-purpose text-to-image human preference reward model. We directly input the image-text pair in the instruction tuning datasets to get the output score.

\subsection{CLIP Score}

CLIP Score~\citep{hessel2021clipscore} is defined as the cosine similarity between the text and image features. The model we used is \textit{clip-vit-large-patch14}.

\subsection{CLIP Features}

We use the \textit{clip-vit-large-patch14} image and text encoders to encode the input samples. The embedding size for both of them is 768, making the final concatenated CLIP feature with a dimension of 1536.

\section{Evaluation Benchmarks and Datasets}
\label{sec:eval}

\paragraph{LVLM-eHub} We use OK-VQA, TextVQA, VisDial, VCR, MSCOCO and Pope in LVLM-eHub for MiniGPT-4. They are all transformed into the visual question answering (VQA) form, and the numbers represent accuracy. Specifically, MCI represents the multi-class identification task, and OC represents the object counting task. Pope~\citep{li2023evaluating} is an evaluation metric for object hallucination.

\paragraph{VLMEvalKit.} We use MMBench, MME, SEED-Bench, Hallusion Bench, Math Vista and ScienceQA on VLMEvalKit for LLaVA experiments. Among them, MMBench, MME and SEED-Bench measure the general ability of VLMs, and a higher number indicates better performance. Hallusion Bench tests the VLMs' robustness against possible language hallucination and visual illusion. \textit{aACC} is the simple average accuracy, \textit{fACC} is the accuracy per figure, and \textit{qACC} is the accuracy per question. More detailed explanations for those metrics can be found in~\citet{liu2023hallusionbench}. Math Vista measures models' math ability, while SiceneQA is the science-related question-answering dataset. Their numbers are the accuracy.

\section{Additional Case study}
\label{sec:casestudy}

In this section, we display the top 5 easiest and hardest samples found by \methodname\ in Table~\ref{tab:top5_easy} and Table~\ref{tab:top5_hard}

\begin{table*}[!htb]
\tiny
\caption{Top 5 Easiest samples.}
\label{tab:top5_easy}
\centering
\begin{tabular}{l|l|l}
\toprule
\textbf{\begin{tabular}[c]{@{}l@{}}1st Easiest\\ Sample\end{tabular}} & \multicolumn{1}{c|}{\textbf{Text}} & \multicolumn{1}{c}{\textbf{Image}} \\ \midrule
\textbf{{[}Human{]}:} & How many people are visible in the image? & \multirow{6}{*}{\includegraphics[width=0.05\linewidth]{figures/000000160938.jpg}} \\
\textbf{{[}GPT{]}:} & There is a man and a woman standing over a table, making a total of two people visible in the image. &  \\
\textbf{{[}Human{]}:} & What is the man doing in the image? &  \\
\textbf{{[}GPT{]}:} & In the image, the man is cutting into a chocolate frosted cake. &  \\
\textbf{{[}Human{]}:} & Are there any other elements on the table besides the cake? &  \\
\textbf{{[}GPT{]}:} & Yes, there are orange flowers on the table in addition to the cake. &  \\ \midrule
\textbf{\begin{tabular}[c]{@{}l@{}}2nd Easiest\\ Sample\end{tabular}} & \multicolumn{1}{c|}{\textbf{Text}} & \multicolumn{1}{c}{\textbf{Image}} \\ \midrule
\textbf{{[}Human{]}:} & Who is sitting at the table in the image? & \multirow{8}{*}{\includegraphics[width=0.1\linewidth]{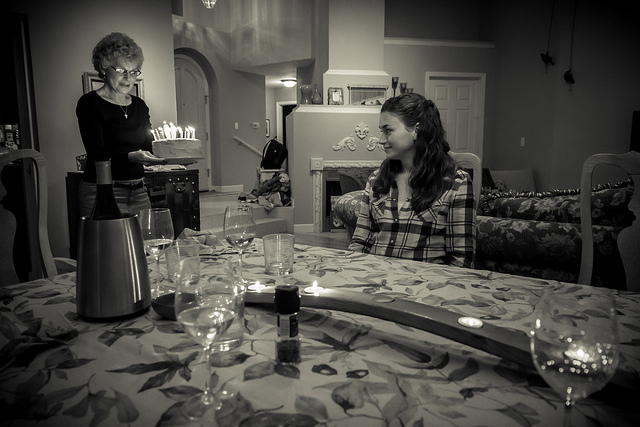}} \\
\textbf{{[}GPT{]}:} & \begin{tabular}[c]{@{}l@{}}A young lady is sitting at the table in the image. She appears to be celebrating her birthday, as there is a birthday cake \\ with lit candles present.\end{tabular} &  \\
\textbf{{[}Human{]}:} & What is happening with the birthday cake? &  \\
\textbf{{[}GPT{]}:} & \begin{tabular}[c]{@{}l@{}}An older woman is holding the birthday cake with lit candles and is bringing it to the young woman who's sitting at \\ the dining room table.\end{tabular} &  \\
\textbf{{[}Human{]}:} & Is there any indication that the photo is old? &  \\
\textbf{{[}GPT{]}:} & \begin{tabular}[c]{@{}l@{}}Based on my visual observation, I cannot determine whether the photo is old or not, as my analysis is purely based on \\ the content of the image and not its quality, color, or other aspects that could suggest its age.\end{tabular} &  \\
\textbf{{[}Human{]}:} & What is the relationship between the young lady and the older woman? &  \\
\textbf{{[}GPT{]}:} & \begin{tabular}[c]{@{}l@{}}Although it is difficult to be certain about the relationship between the young lady and the older woman based on the \\ image alone, the scenario suggests that they may be family members or close friends, as they appear to be sharing a \\ birthday celebration together. The older woman could possibly be the mother, grandmother, or an aunt of the young lady.\end{tabular} &  \\ \midrule
\textbf{\begin{tabular}[c]{@{}l@{}}3rd Easiest\\ Sample\end{tabular}} & \multicolumn{1}{c|}{\textbf{Text}} & \textbf{Image} \\ \midrule
\textbf{{[}Human{]}:} & What activity is the woman in the image engaged in? & \multirow{10}{*}{\includegraphics[width=0.1\linewidth]{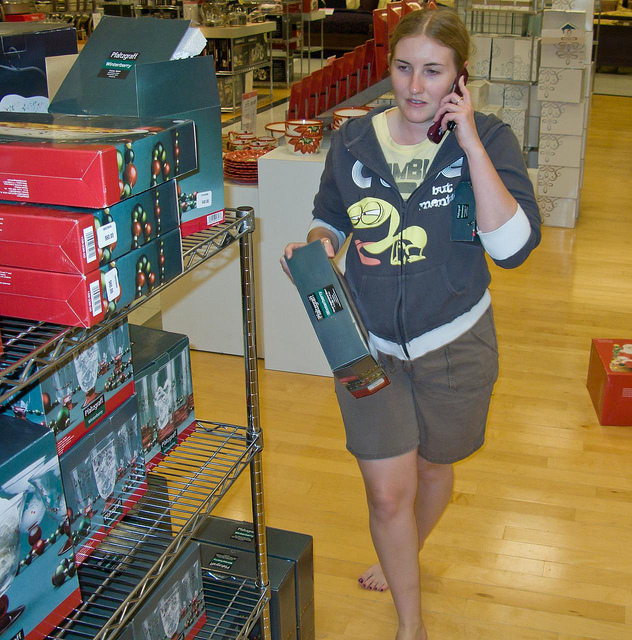}} \\
\textbf{{[}GPT{]}:} & The woman in the image is engaged in shopping while talking on her cell phone inside of a store. &  \\
\textbf{{[}Human{]}:} & What is the woman holding in her hand besides her cell phone? &  \\
\textbf{{[}GPT{]}:} & Besides her cell phone, the woman is holding a small box, which possibly contains a holiday decoration. &  \\
\textbf{{[}Human{]}:} & Is the woman walking or standing still while talking on her phone? &  \\
\textbf{{[}GPT{]}:} & The woman is walking while talking on her cell phone in the store. &  \\
\textbf{{[}Human{]}:} & Are there any noticeable features about the woman's footwear? &  \\
\textbf{{[}GPT{]}:} & The woman is barefooted while talking on her phone, which is an unusual choice of footwear for shopping. &  \\
\textbf{{[}Human{]}:} & What can we infer about the woman's situation during this instance? &  \\
\textbf{{[}GPT{]}:} & \begin{tabular}[c]{@{}l@{}}Based on the available information, we can infer that the woman is multitasking, as she is shopping and talking on her \\ cell phone simultaneously. The fact that she is barefooted suggests she might be preoccupied, in a hurry, or simply \\ relaxed at the moment, not paying particular attention to her choice of footwear. Since she is holding a small box that \\ contains a holiday decoration, it is possible that she is shopping for seasonal items or looking for a gift. The woman is \\ likely in the middle of a discussion or seeking advice from someone on the other end of the phone call, which could be \\ related to choosing the right product or making an important decision regarding her shopping.\end{tabular} &  \\ \midrule
\textbf{\begin{tabular}[c]{@{}l@{}}4th Easiest\\ Sample\end{tabular}} & \multicolumn{1}{c|}{\textbf{Text}} & \textbf{Image} \\ \midrule
\textbf{{[}Human{]}:} & How many women are in the image? & \multirow{10}{*}{\includegraphics[width=0.1\linewidth]{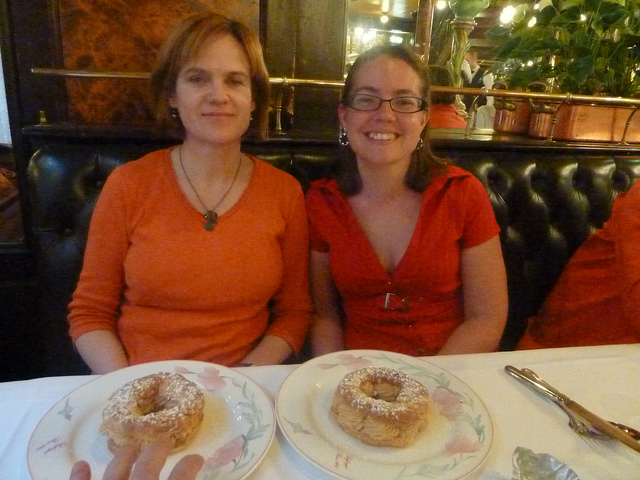}} \\
\textbf{{[}GPT{]}:} & There are two women in the image. &  \\
\textbf{{[}Human{]}:} & What are the two women doing? &  \\
\textbf{{[}GPT{]}:} & The two women are sitting at a table,   having a meal together in a restaurant or a cafe. &  \\
\textbf{{[}Human{]}:} & What type of food is visible on the   table? &  \\
\textbf{{[}GPT{]}:} & There are plates with donuts on the table   in front of the women. &  \\
\textbf{{[}Human{]}:} & What kind of seating arrangement do the   women appear to be in? &  \\
\textbf{{[}GPT{]}:} & The women appear to be sitting in a booth   at a restaurant. &  \\
\textbf{{[}Human{]}:} & Can you describe the social aspect of the   scene? &  \\
\textbf{{[}GPT{]}:} & \begin{tabular}[c]{@{}l@{}}The image portrays a casual and social setting, with two women sharing a meal and enjoying each other's company.   \\ They are likely having a conversation, exchanging stories, or discussing various topics while sitting in the booth. \\ The presence of the donuts on the table suggests that this could be a snack or dessert shared between friends or family \\ members. It reflects a common everyday scenario where people gather around food to strengthen their bonds, relax, \\ and have a good time. The restaurant or cafe setting provides a comfortable and welcoming atmosphere that encourages \\ social interactions and leisurely dining experiences.\end{tabular} &  \\ \midrule
\textbf{\begin{tabular}[c]{@{}l@{}}5th Easiest\\ Sample\end{tabular}} & \multicolumn{1}{c|}{\textbf{Text}} & \textbf{Image} \\ \midrule
\textbf{{[}Human{]}:} & What might be the special occasion the man is celebrating? & \multirow{2}{*}{\includegraphics[width=0.06\linewidth]{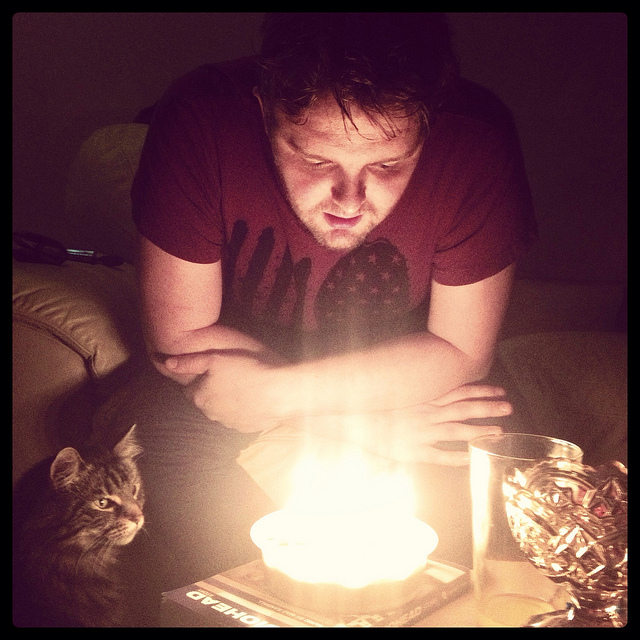}} \\
\textbf{{[}GPT{]}:} & \begin{tabular}[c]{@{}l@{}}Given the image shows a man bending over in front of a cake with lit candles and a cat nearby, it suggests that the   \\ man might be celebrating a special occasion, such as his birthday. The lit candles on the cake are a common tradition \\ during birthday celebrations, and the presence of a cat could indicate that the man is at home with his pet,   enjoying \\ a cozy, intimate celebration.\end{tabular} &  \\ \bottomrule
\end{tabular}
\end{table*}

\begin{table*}[!htb]
\tiny
\caption{Top 5 Hardest samples.}
\label{tab:top5_hard}
\centering
\resizebox{0.8\linewidth}{!}{
\begin{tabular}{l|l|l}
\toprule
\textbf{\begin{tabular}[c]{@{}l@{}}1st Hardest \\ Sample\end{tabular}} & \multicolumn{1}{c|}{\textbf{Text}} & \multicolumn{1}{c}{\textbf{Image}} \\ \hline
\textbf{{[}Human{]}:} & What precautions should someone take when planning to surf in rough waters? & \multirow{2}{*}{\includegraphics[width=0.1\linewidth]{figures/000000129758.jpg}} \\
\textbf{{[}GPT{]}:} & \begin{tabular}[c]{@{}l@{}}When planning to surf in rough waters, like the man wearing a black wetsuit in the image, it is essential to take several\\ precautions to ensure safety. These precautions include:\\ 1. Wearing appropriate gear: A wetsuit provides buoyancy, warmth, and protection from any potential hazards in the \\ water, such as rocks.\\ 2. Checking weather and water conditions: Understanding the current ocean conditions and the forecast will help surfers \\ anticipate how rough the waves might be and decide if it's safe to surf.\\ 3. Being aware of rip currents: These strong currents can carry surfers away from the shore, so it's crucial to understand \\ how to spot and escape them.\\ 4. Surfing with a buddy: Having a surfing partner ensures there's someone to help or alert others in case of an emergency.\\ 5. Knowing personal limits: Understanding one's skill level and not attempting to surf waves that are beyond that level \\ is essential for safety.\\ 6. Proper warm-up and stretching: This can help prevent injuries while surfing in challenging conditions.\\ 7. Communicating with local surfers: They can offer insights into the specific beach break and advise on potential hazards \\ and safe spots.\\ By taking these precautions, surfers can significantly reduce the risks associated with surfing in rough waters and enjoy the \\ exhilaration and challenges that such conditions offer.\end{tabular} &  \\ \hline
\textbf{\begin{tabular}[c]{@{}l@{}}2nd Hardest \\ Sample\end{tabular}} & \multicolumn{1}{c|}{\textbf{Text}} & \multicolumn{1}{c}{\textbf{Image}} \\ \hline
\textbf{{[}Human{]}:} & What challenges do surfers commonly face? & \multirow{2}{*}{\includegraphics[width=0.1\linewidth]{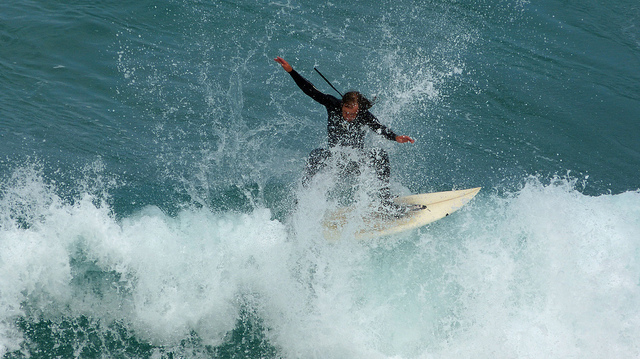}} \\
\textbf{{[}GPT{]}:} & \begin{tabular}[c]{@{}l@{}}Surfers commonly face several challenges related to the dynamic nature of the ocean and the physical demands of the sport. \\ In the image, a person in a wet suit is riding a wave   on a surfboard, which depicts some of these challenges:\\ 1. Variable weather conditions: Surf conditions depend on factors like wind, tides, and swells; these change constantly and \\ might affect the surfer's ability to ride waves safely and effectively.\\ 2. Physical demands: Surfing requires balancing on a surfboard, which demands strong core, upper body, and leg muscles. \\ Paddling to catch waves can be tiring, and surfers need stamina and endurance to stay in the water for extended periods.\\ 3. Timing and technique: Effective surfing requires excellent coordination,   timing, and technique since surfers need to catch \\ waves at the right moment,   stay on the surfboard, and perform various maneuvers, all while maintaining balance.\\ 4. Sharks and other marine creatures: In some regions, surfers may encounter sharks or other potentially dangerous sea creatures. \\ Awareness,   safety measures, and understanding of the marine environment can help minimize the risks associated with such \\ encounters.\\ 5. Rip currents and undertows: These powerful water currents can pose a   risk to surfers as they might make it difficult to \\ return to shore or maintain control in the water.\\ Overall, surfers need to develop their skills, strength, and ocean knowledge to navigate these challenges and enjoy the thrill \\ of riding waves safely.\end{tabular} &  \\ \hline
\textbf{\begin{tabular}[c]{@{}l@{}}3rd Hardest\\ Sample\end{tabular}} & \multicolumn{1}{c|}{\textbf{Text}} & \multicolumn{1}{c}{\textbf{Image}} \\ \hline
\textbf{{[}Human{]}:} & What are some factors that can affect this surfer's performance and safety? & \multirow{2}{*}{\includegraphics[width=0.1\linewidth]{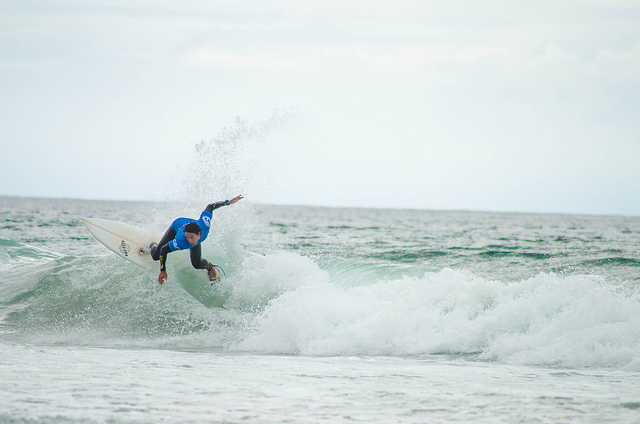}} \\
\textbf{{[}GPT{]}:} & \begin{tabular}[c]{@{}l@{}}Several factors can affect this surfer's performance and safety while riding the wave. Some of these factors include:\\ 1. Wave conditions: The size, power, and shape of the wave can influence the surfer's ability to perform maneuvers and maintain \\ balance. Unexpected changes in wave conditions or encountering a challenging wave could increase the risk of falling, getting \\ caught inside or even being pushed towards hazardous areas.\\ 2. Surfer's skill level: The surfer's experience and skill level play a crucial role in their performance and safety. A less experienced \\ surfer might struggle to maintain control and execute complex maneuvers, which could result in accidents or injuries.\\ 3. Equipment: The surfboard, leash, and wetsuit can impact the surfer's performance and safety. A poorly chosen surfboard might \\ make it challenging to navigate the waves effectively, and a damaged leash could cause the surfer to lose their board in a wipeout, \\ increasing the risk of injury.\\ 4. Weather conditions: Wind and tide conditions can affect the surf and the surfer's ability to ride the wave smoothly. Strong winds \\ might make it difficult to maintain balance and control, while extreme tide conditions may expose hidden obstacles or create \\ unfavorable surfing conditions.\\ 5. Ocean environment: Hidden rocks, reefs, or other hazards in the water can pose risks to the surfer's safety. Additionally, ocean \\ currents and rip tides can move the surfer away from their intended position, which could lead to dangerous situations.\\ By being aware of these factors, surfers can take necessary precautions,   improve their skills, and choose suitable equipment to \\ ensure a safe and enjoyable surfing experience.\end{tabular} &  \\ \hline
\textbf{\begin{tabular}[c]{@{}l@{}}4th Hardest\\ Sample\end{tabular}} & \multicolumn{1}{c|}{\textbf{Text}} & \multicolumn{1}{c}{\textbf{Image}} \\ \hline
\textbf{{[}Human{]}:} & How many people are surfing in the image? & \multirow{8}{*}{\includegraphics[width=0.1\linewidth]{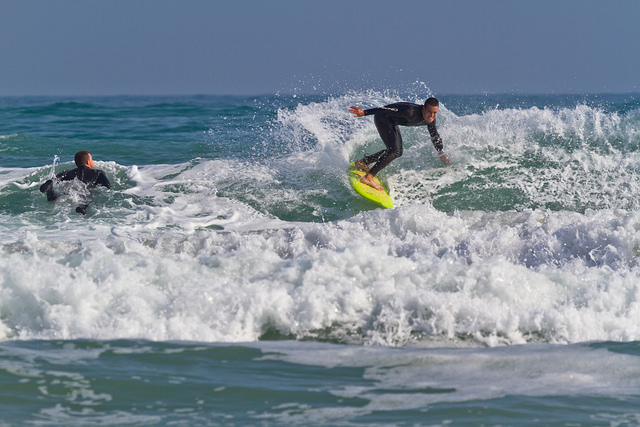}} \\
\textbf{{[}GPT{]}:} & There are two people surfing in the image. &  \\
\textbf{{[}Human{]}:} & What is the condition of the waves in the ocean? &  \\
\textbf{{[}GPT{]}:} & The waves in the ocean are big and choppy, providing an exciting and challenging surfing experience for the surfers. &  \\
\textbf{{[}Human{]}:} & Is one of the surfers standing up on the surfboard? &  \\
\textbf{{[}GPT{]}:} & Yes, one of the surfers is standing up on the surfboard, while the other surfer is not. &  \\
\textbf{{[}Human{]}:} & What are some challenges surfers might face when riding big, choppy waves? &  \\
\textbf{{[}GPT{]}:} & \begin{tabular}[c]{@{}l@{}}Riding big, choppy waves presents several challenges for surfers, some of which include:\\ 1. Balance and stability: Maintaining balance on a surfboard while maneuvering choppy waves can be difficult as the \\ unpredictable movement of the water makes it harder for surfers to find a steady footing.\\ 2. Timing: Catching and riding a big wave requires precise timing. Surfers need to paddle at the right speed and position themselves \\ properly to catch the wave. Choppy conditions can make it harder to determine the best moment to catch a wave and ride it successfully.\\ 3. Strength and endurance: To successfully ride choppy waves, surfers must have strong upper body and leg muscles to paddle, maintain \\ balance, and push through the turbulence. These waves can also be physically exhausting, which requires good overall fitness and endurance.\\ 4. Risk of accidents: Big, choppy waves increase the risk of accidents like wipeouts, collisions with other surfers, and encounters with \\ dangerous underwater obstacles. The rough conditions might also cause surfers to lose control of their boards, potentially resulting in injuries.\\ 5. Unpredictability: The unpredictable nature of choppy waves makes it more challenging for surfers to anticipate the wave's behavior \\ and adjust their techniques accordingly. Surfers must be highly adaptable and capable of making split-second decisions to maneuver\\  through these conditions effectively.\\ In conclusion, surfing big, choppy waves requires advanced skills, good physical fitness, attentiveness, and adaptability. The unpredictable \\ nature of the water and the potential risks involved make it a thrilling yet challenging activity for surfers.\end{tabular} &  \\ \hline
\textbf{\begin{tabular}[c]{@{}l@{}}5th Hardest\\ Sample\end{tabular}} & \multicolumn{1}{c|}{\textbf{Text}} & \multicolumn{1}{c}{\textbf{Image}} \\ \hline
\textbf{{[}Human{]}:} & How many surfers are visible in the image? & \multirow{10}{*}{\includegraphics[width=0.1\linewidth]{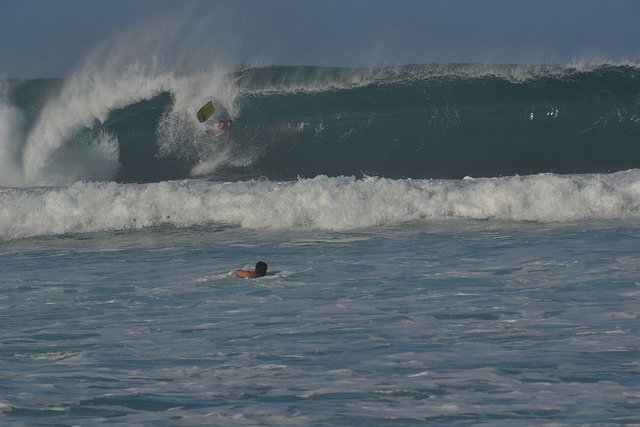}} \\
\textbf{{[}GPT{]}:} & There are two surfers visible in the image - one riding a wave on top of a surfboard, and the other swimming out to surf. &  \\
\textbf{{[}Human{]}:} & What is happening to the surfer on the surfboard? &  \\
\textbf{{[}GPT{]}:} & The surfer on the surfboard is in the midst of a fall as the wave is breaking, resulting in a wipeout. &  \\
\textbf{{[}Human{]}:} & What is the second person doing in the water? &  \\
\textbf{{[}GPT{]}:} & The second person is swimming out towards the waves, possibly intending to surf or catch some waves as well. &  \\
\textbf{{[}Human{]}:} & What is the overall atmosphere of the scene in the image? &  \\
\textbf{{[}GPT{]}:} & \begin{tabular}[c]{@{}l@{}}The overall atmosphere of the scene in the image is dynamic and action-packed, with surfers engaging in their sport amidst the powerful \\ waves. It captures the excitement and challenges of surfing, as well as the skill and determination required to ride the waves successfully.\end{tabular} &  \\
\textbf{{[}Human{]}:} & What are the potential risks of wipeouts in surfing, and how can surfers minimize these risks? &  \\
\textbf{{[}GPT{]}:} & \begin{tabular}[c]{@{}l@{}}Wipeouts are an inherent part of surfing and can pose risks to surfers due to the potential for injury or more serious accidents. Some of \\ the potential risks associated with wipeouts include:\\ 1. Collisions with the surfboard or other objects: Falling off the surfboard can result in surfers colliding with their own boards, causing \\ injuries. Additionally, they might collide with rocks, reefs, or other surfers in crowded surf breaks.\\ 2. Drowning or near-drowning: Wipeouts can cause surfers to be held underwater by the wave, resulting in disorientation and \\ potentially drowning if they are unable to resurface in time.\\ 3. Being caught in a rip current: A wipeout may force the surfer into a rip current, a strong water stream that can carry them away\\  from the shore quickly and create difficulty in returning to the beach.\\ 4. Impact injuries: A hard fall from a surfboard can result in the surfer hitting the water surface with force, leading to injuries \\ such as sprains,   fractures, or even head trauma.\\ To minimize these risks, surfers can take several precautions:\\ 1. Wearing appropriate protective gear: Using helmets, wetsuits or rash guards, and other protective equipment can help shield the\\  surfer's body from injuries during wipeouts.\\ 2. Developing swimming and water safety skills: Surfers should ensure they are strong swimmers and understand ocean safety\\ principles, such as how to identify and escape from rip currents and using the best techniques for resurfacing after being held underwater.\\ 3. Practicing proper surfing techniques: Mastering the sport's fundamentals, including paddling, duck diving, and foot positioning on \\ the board, can help surfers maintain control and minimize the risk of wipeouts.\\ 4. Surfing within their skill level: Surfers should be aware of their abilities and choose waves, surf breaks, and conditions that match\\  their skill level, avoiding situations that might expose them to unnecessary risk.\\ 5. Staying aware of their surroundings: Surfers should always maintain situational awareness in the water, paying attention to other\\  surfers, potential hazards, and changing ocean conditions to prevent accidents.\\ By following these recommendations, surfers can reduce the risks associated with wipeouts and enjoy the thrilling experience\\  of surfing more safely.\end{tabular} &  \\ 
\bottomrule
\end{tabular}
}
\end{table*}

\section{MiniGPT4 Experimental Results}
\label{sec:minigpt4_number}

We provide the exact numbers for the main experimental results on MiniGPT-4 in Table~\ref{tab:minigpt4_main}.

\begin{table*}[!htb]
\caption{Main results on MiniGPT-4.}
\label{tab:minigpt4_main}
\centering
\begin{tabular}{c|cccc}
\toprule
\textbf{}             & \multicolumn{1}{l}{\textbf{Full Data (3.4k)}} & \multicolumn{1}{l}{\textbf{Random (500)}} & \multicolumn{1}{l}{\textbf{w/o Diversity (500)}} & \multicolumn{1}{l}{\textbf{Ours (500)}} \\ \midrule
\textbf{OK-VQA}       & \textbf{40.07}                                & 35.08                                     & 36.86                                            & 38.74                                   \\
\textbf{TextVQA}      & 21.36                                         & 20.02                                     & 21.94                                            & \textbf{21.96}                          \\
\textbf{VisDial}      & \textbf{66.74}                                & 55.35                                     & 60.75                                            & 65.39                                   \\
\textbf{VCR1\_MCI}    & 54.56                                         & 46.26                                     & 58.18                                            & \textbf{60.01}                          \\
\textbf{VCR1\_OC}     & 34.56                                         & 32.39                                     & \textbf{36.93}                                   & 36.80                                   \\
\textbf{MSCOCO\_MCI}  & 48.67                                         & 42.32                                     & 51.48                                            & \textbf{51.68}                          \\
\textbf{MSCOCO\_OC}   & 37.71                                         & 36.44                                     & \textbf{40.81}                                   & 40.63                                   \\
\textbf{MSCOCO\_Pope} & 54.34                                         & 52.85                                     & 56.56                                            & \textbf{58.63}                          \\ \midrule
\textbf{Average}      & 44.75                                         & 40.09                                     & 45.44                                            & \textbf{46.73}                          \\ \bottomrule
\end{tabular}
\end{table*}

\end{document}